\documentclass[letterpaper, 10 pt, conference]{ieeeconf}

\IEEEoverridecommandlockouts

\usepackage{multirow}
\usepackage{graphicx}
\usepackage{comment}
\usepackage{cite}
\usepackage{amsmath}
\usepackage{amsfonts}
\usepackage{mathtools}
 \usepackage{tabularx}
\usepackage{graphicx}
\usepackage{array}
\usepackage{float}
\usepackage{hhline}
\usepackage{algorithmic}
\usepackage[linesnumbered, ruled, vlined]{algorithm2e}
\usepackage[colorlinks,urlcolor=blue]{hyperref}
\usepackage{setspace}
\usepackage{colortbl}
\usepackage{arydshln}
\SetCommentSty{mycommfont}

\DeclareMathOperator*{\argmin}{arg\,min}

\title{\LARGE\bf Direct LiDAR-Inertial Odometry: \\ 
Lightweight LIO with Continuous-Time Motion Correction\vspace{-4mm}}

\author{Kenny Chen$^{1}$, Ryan Nemiroff$^{1}$, and Brett T. Lopez$^{2}$%
\thanks{$^{1}$Kenny Chen and Ryan Nemiroff are with the Department of Electrical and Computer Engineering, University of California Los Angeles, Los Angeles, CA, USA. {\tt\footnotesize \{kennyjchen, ryguyn\}@ucla.edu}}%
\thanks{$^{2}$Brett T. Lopez is with the Department of Mechanical and Aerospace Engineering, University of California Los Angeles, Los Angeles, CA, USA. {\tt\footnotesize btlopez@ucla.edu}}%
\thanks{All authors are with the Verifiable and Control-Theoretic Robotics Laboratory, University of California Los Angeles, Los Angeles,
CA, USA.}%
}

\begin{document}
\IEEEaftertitletext{\vspace{-2mm}}
\maketitle
\thispagestyle{empty}
\pagestyle{empty}


\begin{abstract}

Aggressive motions from agile flights or traversing irregular terrain induce motion distortion in LiDAR scans that can degrade state estimation and mapping. Some methods exist to mitigate this effect, but they are still too simplistic or computationally costly for resource-constrained mobile robots. To this end, this paper presents Direct LiDAR-Inertial Odometry (DLIO), a lightweight LiDAR-inertial odometry algorithm with a new coarse-to-fine approach in constructing continuous-time trajectories for precise motion correction. The key to our method lies in the construction of a set of analytical equations which are parameterized solely by time, enabling fast and parallelizable point-wise deskewing. This method is feasible only because of the strong convergence properties in our nonlinear geometric observer, which provides provably correct state estimates for initializing the sensitive IMU integration step. Moreover, by simultaneously performing motion correction and prior generation, and by directly registering each scan to the map and bypassing scan-to-scan, DLIO's condensed architecture is nearly 20\% more computationally efficient than the current state-of-the-art with a 12\% increase in accuracy. We demonstrate DLIO's superior localization accuracy, map quality, and lower computational overhead as compared to four state-of-the-art algorithms through extensive tests using multiple public benchmark and self-collected datasets.


\end{abstract}


\section{Introduction}

Accurate real-time state estimation and mapping are necessary capabilities for mobile robots to perceive, plan, and navigate through unknown environments. LiDAR-based localization has recently become a viable option for many mobile platforms, such as drones, due to more compact and accurate sensors. As a result, researchers have developed several new LiDAR odometry (LO) and LiDAR-inertial odometry (LIO) algorithms which often outperform vision-based approaches due to the superior range and depth measurement accuracy of a LiDAR. However, there are still fundamental challenges in developing reliable and accurate LiDAR-centric algorithms \cite{cadena2016past}, especially for robots that execute agile maneuvers or traverse uneven terrain. In particular, such aggressive movements can induce significant distortion in the point cloud which corrupts the scan-matching process, resulting in severe or catastrophic localization error and map deformation. 

Existing algorithms which attempt to compensate for this effect may work well in structured environments for non-holonomic systems (e.g., autonomous driving), but their performance can degrade under irregular conditions due to simplistic motion models, loss in precision from discretization, and/or computational inefficiencies. For instance, works such as \cite{zhang2014loam, shan2018lego, shan2020lio} assume constant velocity during scan acquisition which may work well for simple, predictable trajectories, but this quickly breaks down under significant acceleration. On the other hand, \cite{xu2021fast} and \cite{xu2022fast} use a back-propagation technique to mitigate distortion for each point, but their method may induce a loss in precision from accumulating integration error over time. More recently, continuous-time methods attempt to fit a smooth trajectory over a set of control points \cite{park2018elastic, ramezani2022wildcat} or augment scan-matching optimization with additional free variables \cite{dellenbach2022ct}, but such methods still hold strong assumptions on the trajectory (i.e., smooth movement) or may be too computationally costly for weight-limited platforms.

\begin{figure}[!t]
    \centering
    \vspace{2mm}
    \includegraphics[width=0.99\columnwidth]{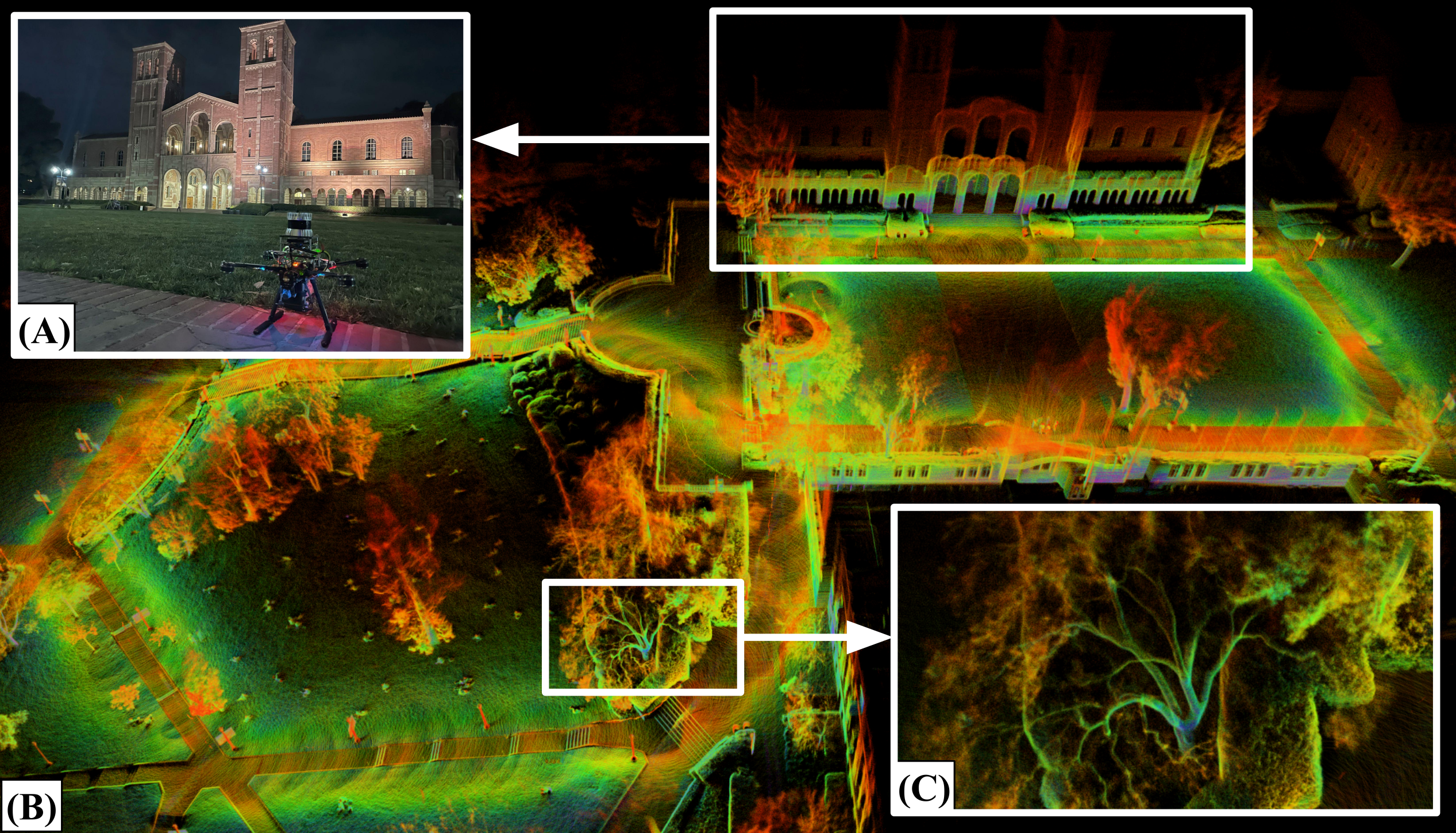}
    \vspace{-6mm}
    \caption{\textbf{Real-time Localization and Dense Mapping.} DLIO generates detailed maps by reliably estimating robot pose, velocity, and sensor biases in real-time. (A) Our custom aerial vehicle next to UCLA's Royce Hall. (B) A bird's eye view of Royce Hall and its surroundings generated by DLIO. (C) A close-up of a tree, showcasing the fine detail that DLIO is able to capture in its output map. Color denotes intensity of point return.}
    \vspace{-1mm}
    \label{fig:main}
\end{figure}

\begin{figure*}[!t]
    \centering
    \vspace{2mm}
    \includegraphics[width=0.95\textwidth]{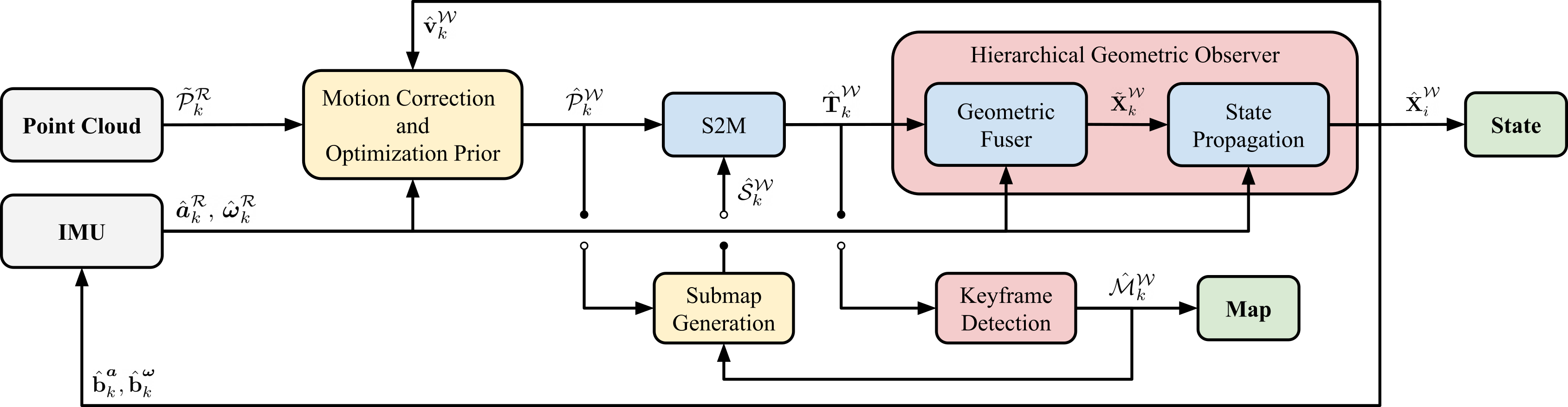}
    \vspace{-2mm}
    \caption{\textbf{System Architecture.} DLIO's lightweight architecture combines motion correction and prior construction into a single step, in addition to removing the scan-to-scan module previously required for LiDAR-based odometry. Point-wise continuous-time integration in $\mathcal{W}$ ensures maximum fidelity of the corrected cloud and is registered onto the robot's map by a custom GICP-based scan-matcher. The system's state is subsequently updated by a nonlinear geometric observer with strong convergence properties \cite{lopez2023contracting}, and these estimates of pose, velocity, and bias then initialize the next iteration.}
    \label{fig:architecture}
    \vspace{-5mm}
\end{figure*}

To this end, we present Direct LiDAR-Inertial Odometry (DLIO), a fast and reliable odometry algorithm that provides accurate localization and detailed 3D mapping (Fig.~\ref{fig:main}) with four main contributions. First, we propose a new coarse-to-fine technique for constructing continuous-time trajectories, in which a set of analytical equations with a constant jerk and angular acceleration motion model is derived for fast and parallelizable point-wise motion correction. Second, a novel condensed architecture is presented which combines motion correction and prior construction into one step and directly performs scan-to-map registration, significantly reducing overall computational overhead of the algorithm. 
Third, we leverage a new nonlinear geometric observer \cite{lopez2023contracting} that possesses strong performance guarantees---critical for achieving the first two contributions---in the pipeline to robustly generate accurate estimates of the robot's full state with minimal computational complexity.
Finally, the efficacy of our approach is verified through extensive experimental results using multiple datasets against the state-of-the-art.


\section{Related Work}

Geometric LiDAR odometry algorithms rely on aligning point clouds by solving a nonlinear least-squares problem that minimizes the error across corresponding points and/or planes. To find point/plane correspondences, methods such as the iterative closest point (ICP) algorithm \cite{besl1992method, chen1992object} or Generalized-ICP (GICP) \cite{segal2009generalized} recursively match entities until alignment converges to a local minimum. 
Slow convergence time is often observed when determining correspondences for a large set of points, so 
\textit{feature}-based methods \cite{zhang2014loam, shan2018lego, shan2020lio, shan2021lvi, pan2021mulls, xu2021fast, nguyen2021miliom, ye2019tightly} attempt to extract only the most salient data points, e.g., corners and edges, in a scan to decrease computation time. However, useful points are often discarded as the efficacy of feature extraction is highly dependent on specific implementation. Conversely, \textit{dense} methods \cite{palieri2020locus, tagliabue2020lion, chen2022direct, xu2022fast, reinke2022iros} directly align acquired scans but often rely heavily on aggressive voxelization---a process that can alter important data correspondences---to achieve real-time performance.

LiDAR odometry approaches can also be broadly classified according to their method of incorporating other sensing modalities into the estimation pipeline.
\textit{Loosely}-coupled methods \cite{zhang2014loam, shan2018lego, palieri2020locus, tagliabue2020lion, chen2022direct} process data sequentially.
For example, IMU measurements are used to augment LiDAR scan registration by providing an optimization prior. 
These methods are often quite robust due to the precision of LiDAR measurements, but localization results can be less accurate as only a subset of all available data is used for estimation.
\textit{Tightly}-coupled methods \cite{shan2020lio, xu2022fast, ye2019tightly, nguyen2021miliom}, on the otherhand, can offer improved accuracy by jointly considering measurements from all sensing modalities.
These methods commonly employ either graph-based optimization \cite{shan2020lio, ye2019tightly, nguyen2021miliom, zhang2016degeneracy} or a stochastic filtering framework, e.g., Kalman filter \cite{xu2021fast, xu2022fast}. However, compared to geometric observers \cite{baldwin2007complementary, vasconcelos2008nonlinear}, these approaches possess minimal convergence guarantees even in the most ideal settings which can result in significant localization error from inconsistent sensor fusion and map deformation from incorrect scan placement.

Incorporating additional sensors can also aid in correcting motion-induced point cloud distortion. 
For example, LOAM~\cite{zhang2014loam} compensates for spin distortion by iteratively estimating sensor pose via scan-matching and a loosely-coupled IMU using a constant velocity assumption. Similarly, LIO-SAM~\cite{shan2020lio} formulates LiDAR-inertial odometry atop a factor graph to jointly optimize for body velocity, and in their implementation, points were subsequently deskewed by linearly interpolating rotational motion. FAST-LIO~\cite{xu2021fast} and FAST-LIO2~\cite{xu2022fast} instead employ a back-propagation step on point timestamps after a forward-propagation of IMU measurements to produce relative transformations to the scan-end time. However, these methods (and others \cite{renzler2020increased, deschenes2021lidar}) all operate in \textit{discrete}-time which may induce a loss in precision, leading to a high interest in \textit{continuous}-time methods. Elastic LiDAR Fusion~\cite{park2018elastic}, for example, handles scan deformation by optimizing for a continuous linear trajectory, whereas Wildcat~\cite{ramezani2022wildcat} and \cite{droeschel2018efficient} instead iteratively fit a cubic B-spline to remove distortion from surfel maps. More recently, CT-ICP~\cite{dellenbach2022ct} and ElasticLiDAR++~\cite{park2022elasticity} use a LiDAR-only approach to define a continuous-time trajectory parameterized by two poses per scan, which allows for elastic registration of the scan during optimization. However, these methods can still be too simplistic in modeling the trajectory under highly dynamical movements or may be too computationally costly to work reliably in real-time.

To this end, DLIO proposes a fast, coarse-to-fine approach to construct each inter-sweep trajectory for accurate motion correction. A discrete set of poses is first computed via numerical integration on IMU measurements, and smooth trajectories between measurement samples are subsequently built via analytical, continuous-time equations to query each unique per-point deskewing transform. Our approach is fast in that we solve a set of analytical equations rather than an optimization problem (e.g., spline-fitting), parameterized solely by the timestamp of the point which can be easily parallelized. Our approach is also accurate in that we use a higher-order motion model to represent the underlying system dynamics which can capture high-frequency movements that may otherwise be lost in methods that attempt to fit a smooth trajectory to a set of control points. This approach is built into a simplified LIO architecture which performs motion correction and GICP prior construction in one shot, in addition to performing scan-to-map alignment directly without the intermediary scan-to-scan; this is all possible through the strong convergence guarantees of our novel geometric observer with provably correct state estimates.


\section{Method}

\subsection{System Overview}

DLIO is a lightweight LIO algorithm that generates robot state estimates and geometric maps through a unique architecture that contains two main components with three innovations (Fig.~\ref{fig:architecture}). The first is a fast scan-matcher which registers dense, motion-corrected point clouds onto the robot's map by performing alignment with an extracted local submap. Point-wise continuous-time integration in $\mathcal{W}$ ensures maximum image fidelity of the corrected cloud while simultaneously building in a prior for GICP optimization. In the second, a nonlinear geometric observer \cite{lopez2023contracting} updates the system's state with the first component's pose output to provide high-rate and provably correct estimates of pose, velocity, and sensor biases which converge globally. These estimates then initialize the next iteration of motion correction, scan-matching, and state update.

\subsection{Notation}

Let the point cloud for a single LiDAR sweep initiated at time $t_k$ be denoted as $\mathcal{P}_k$ and indexed by $k$.
The point cloud $\mathcal{P}_k$ is composed of points $p^n_{k} \in \mathbb{R}^3$ that are measured at a time $\Delta t^n_{k}$ relative to the start of the scan and indexed by ${n=1,\hdots,N}$ where $N$ is the total number of points in the scan.
The world frame is denoted as $\mathcal{W}$ and the robot frame as $\mathcal{R}$ located at its center of gravity, with the convention that $x$ points forward, $y$ left, and $z$ up. The IMU's coordinate system is denoted as $\mathcal{B}$ and the LiDAR's as $\mathcal{L}$, and the robot's state vector $\textbf{X}_k$ at index $k$ is defined as the tuple
\begin{equation}
    \label{eq:state}
    \textbf{X}_k = \left[ \, \textbf{p}^{\mathcal{W}}_{k} ,\, \textbf{q}^{\mathcal{W}}_{k} ,\, \textbf{v}^{\mathcal{W}}_{k} ,\, \textbf{b}^{\boldsymbol{a}}_{k} ,\, \textbf{b}^{\boldsymbol{\omega}}_{k} \, \right] ^\top \,,
\end{equation}
\noindent where $\textbf{p}^{\mathcal{W}} \in \mathbb{R}^3$ is the robot's position, $\textbf{q}^{\mathcal{W}}$ is the orientation encoded by a four vector quaternion on $\mathbb{S}^3$ under Hamilton notation, $\textbf{v}^{\mathcal{W}} \in \mathbb{R}^3$ is the robot's velocity, $\textbf{b}^{\boldsymbol{a}} \in \mathbb{R}^3$ is the accelerometer's bias, and $\textbf{b}^{\boldsymbol{\omega}} \in \mathbb{R}^3$ is the gyroscope's bias. Measurements $\hat{\boldsymbol{a}}$ and $\hat{\boldsymbol{\omega}}$ from an IMU are modeled as
\begin{align}
    \label{eq:imu}
    \hat{\boldsymbol{a}}_i &= (\boldsymbol{a}_i - \boldsymbol{g}) + \textbf{b}_i^{\boldsymbol{a}} + \textbf{n}_i^{\boldsymbol{a}} \,, \\
    \hat{\boldsymbol{\omega}}_i &= \boldsymbol{\omega}_i + \textbf{b}_i^{\boldsymbol{\omega}} + \textbf{n}_i^{\boldsymbol{\omega}} \,,
\end{align}
\noindent and indexed by ${i=1,\hdots,M}$ for $M$ measurements between clock times $t_{k\text{-}1}$ and $t_{k}$. With some abuse of notation, indices $k$ and $i$ occur at LiDAR and IMU rate, respectively, and will be written this way for simplicity unless otherwise stated. Raw sensor measurements $\boldsymbol{a}_i$ and $\boldsymbol{\omega}_i$ contain bias $\textbf{b}_i$ and white noise $\textbf{n}_i$, and $\boldsymbol{g}$ is the rotated gravity vector. In this work, we address the following problem: given an accumulated point cloud $\mathcal{P}_k$ from a LiDAR and measurements $\boldsymbol{a}_{i}$ and $\boldsymbol{\omega}_{i}$ sampled between each received scan by an IMU, estimate the robot's state $\hat{\textbf{X}}_i^{\mathcal{W}}$ and the geometric map $\hat{\mathcal{M}}_k^{\mathcal{W}}$.

\begin{algorithm}[!tb]
    \small
    \setstretch{1}
	\SetAlgoLined
	\textbf{input:} $\hat{\textbf{X}}_{k\text{-}1}^{\mathcal{W}}$, $\mathcal{P}_k^{\mathcal{L}}$, $\boldsymbol{a}_{k}^{\mathcal{B}}$,  $\boldsymbol{\omega}_{k}^{\mathcal{B}}$ ; \,
	\textbf{output:} $\hat{\textbf{X}}_i^{\mathcal{W}}$, $\hat{\mathcal{M}}_k^{\mathcal{W}}$ \\
	
	\BlankLine
	
	\small {\tcp{LiDAR Callback Thread}}
	\While{$\mathcal{P}_k^{\mathcal{L}} \neq \emptyset$} {
	    \small {\tcp{initialize points and transform to $\mathcal{R}$}}
	    $\mathcal{\tilde{P}}_k^\mathcal{R}$ $\leftarrow$ initializePointCloud$(\,\mathcal{P}_k^\mathcal{L}\,)$ (Sec.~\ref{sec:preprocessing});\\
	    \small {\tcp{continuous-time motion correction}}
	    \For{$\hat{\boldsymbol{a}}_i^{\mathcal{R}}, \hat{\boldsymbol{\omega}}_i^{\mathcal{R}}$ between $t_{k\text{-}1}$ and $t_{k}$} {
	        $\hat{\textbf{p}}_{i}, \hat{\textbf{v}}_{i}, \hat{\textbf{q}}_{i} \leftarrow$ discreteInt$(\,\hat{\textbf{X}}_{k\text{-}1}^{\mathcal{W}},\,\hat{\boldsymbol{a}}_{i\text{-}1}^{\mathcal{R}} ,\, \hat{\boldsymbol{\omega}}_{i\text{-}1}^{\mathcal{R}}\,) \, (\ref{eq:deskew})$; \\ $\hat{\textbf{T}}_{i}^{\mathcal{W}} = [ \, \hat{\textbf{R}}(\hat{\textbf{q}}_{i}) \, | \, \hat{\textbf{p}}_{i} \, ]$;\\
	    }
	    \For{$p_k^n \in \mathcal{\tilde{P}}_k^\mathcal{R}$} {
	        $\hat{\textbf{T}}^{\mathcal{W}*}_{n} \leftarrow$ continuousInt$(\,\hat{\textbf{T}}^{\mathcal{W}*}_{i},\, t_n\,) \,\, (\ref{eq:deskew_timestamp})$; \\
	        $\hat{p}_k^n = \hat{\textbf{T}}^{\mathcal{W}*}_{n} \otimes p_k^n \,$; 
	        $\hat{\mathcal{P}}_k^\mathcal{W}$.append$(\,\hat{p}_k^n\,)$;\\
	    }
	    \small {\tcp{scan-to-map registration}}
	    $\hat{\mathcal{S}}_k^\mathcal{W} \leftarrow$ generateSubmap$(\,\hat{\mathcal{M}}_k^{\mathcal{W}}\,)$ \cite{chen2022direct}\\
	    $\hat{\textbf{T}}_k^{\mathcal{W}} \leftarrow$ GICP$(\,\hat{\mathcal{P}}_k^\mathcal{W},\, \hat{\mathcal{S}}_k^\mathcal{W}\,) \,$ (\ref{eq:gicp});\\
	    \small {\tcp{geometric observer: state update}}
	    $\hat{\textbf{X}}_k^\mathcal{W} \leftarrow$ update$(\,\hat{\textbf{T}}_k^{\mathcal{W}},\, \Delta t^{+}_k\,) \,$ (Sec.~\ref{methods:geo});\\
	    \small {\tcp{update keyframe map}}
	    \lIf{$\hat{\mathcal{P}}_k^\mathcal{W}$ is a keyframe} {
            $\hat{\mathcal{M}}_k^{\mathcal{W}} \leftarrow \hat{\mathcal{M}}_{k\text{-}1}^{\mathcal{W}} \oplus \hat{\mathcal{P}}_k^\mathcal{W}$
	    }
	    \Return $\hat{\textbf{X}}_k^{\mathcal{W}},\, \hat{\mathcal{M}}_k^{\mathcal{W}}$ \\
	    
	}
	
	\BlankLine
	
	\small {\tcp{IMU Callback Thread}}
	\While{$\boldsymbol{a}_i^{\mathcal{B}} \neq \emptyset$ and $\boldsymbol{\omega}_i^{\mathcal{B}} \neq \emptyset$} {
	    \small {\tcp{apply biases and transform to $\mathcal{R}$}}
	    $\hat{\boldsymbol{a}}_i^{\mathcal{R}} ,\, \hat{\boldsymbol{\omega}}_i^{\mathcal{R}}$ $\leftarrow$ initializeImu$(\,\boldsymbol{a}_i^{\mathcal{B}} ,\, \boldsymbol{\omega}_i^{\mathcal{B}}\,)$ (Sec.~\ref{sec:preprocessing}); \\
	    \small {\tcp{geometric observer: state propagation}}
	    $\hat{\textbf{X}}_i^{\mathcal{W}} \leftarrow$ propagate$(\,\hat{\textbf{X}}_{k}^{\mathcal{W}},\, \hat{\boldsymbol{a}}_{i}^{\mathcal{R}},\, \hat{\boldsymbol{\omega}}_{i}^{\mathcal{R}},\, \Delta t^{+}_i\,)$ (Sec.~\ref{methods:geo});\\
	    \Return $\hat{\textbf{X}}_i^{\mathcal{W}}$ \\
	    
	}

	\caption{Direct LiDAR-Inertial Odometry}
	\label{alg:dlio}
	
\end{algorithm}

\subsection{Preprocessing}
\label{sec:preprocessing}

The inputs to DLIO are a dense 3D point cloud collected by a modern 360$^\circ$ mechanical LiDAR, such as an Ouster or a Velodyne (10-20Hz), in addition to time-synchronized linear acceleration and angular velocity measurements from a 6-axis IMU at a much higher rate (100-500Hz). Prior to downstream tasks, all sensor data is transformed to be in $\mathcal{R}$ located at the robot's center of gravity via extrinsic calibration. For IMU, effects of displacing linear acceleration measurements on a rigid body must be considered if the sensor is not coincident with the center of gravity; this is done by considering all contributions of linear acceleration at $\mathcal{R}$ via the cross product between angular velocity and the offset of the IMU. To minimize information loss, we do not preprocess the point cloud except for a box filter of size 1m$^3$ around the origin to remove points that may be from the robot itself, and a light voxel filter for higher resolution clouds. This distinguishes our work from others that either attempt to detect features (e.g., corners, edges, or surfels) or heavily downsamples the cloud through a voxel filter.

\subsection{Continuous-Time Motion Correction with Joint Prior}
\label{methods:motion_correction_and_prior}

\begin{figure}[!t]
    \centering
    \vspace{2mm}
    \includegraphics[width=0.85\columnwidth]{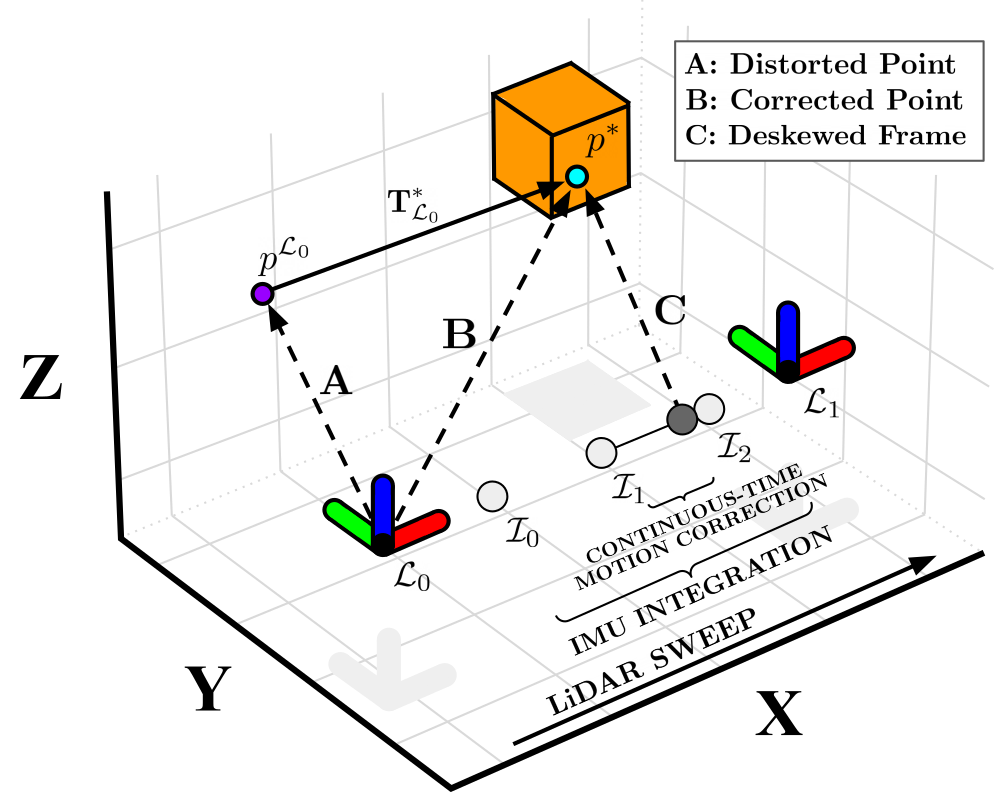}
    \vspace{-2mm}
    \caption{\textbf{Coarse-to-Fine Point Cloud Deskewing.} A distorted point $p^{\mathcal{L}_0}$ (A) is deskewed through a two-step process which first integrates IMU measurements between scans, then solves for a unique transform in continuous-time (C) for the original point which deskews $p^{\mathcal{L}_0}$ to $p^*$ (B).}
    \label{fig:motion_correction}
\end{figure}

Point clouds from spinning LiDAR sensors suffer from motion distortion during movement due to the rotating laser array collecting points at different instances during a sweep. Rather than assuming simple motion (i.e., constant velocity) during sweep that may not accurately capture fine movement, we instead use a more accurate constant jerk and angular acceleration model to compute a unique transform for each point via a two-step coarse-to-fine propagation scheme. This strategy aims to minimize the errors that arise due to the sampling rate of the IMU and the time offset between IMU and LiDAR point measurements. 
Trajectory throughout a sweep is first coarsely constructed through numerical IMU integration \cite{forster2016manifold}, which is subsequently refined by solving a set of analytical continuous-time equations in $\mathcal{W}$ (Fig.~\ref{fig:motion_correction}).

\begin{figure}[!t]
    \centering
    \vspace{2mm}
    \includegraphics[width=0.85\columnwidth]{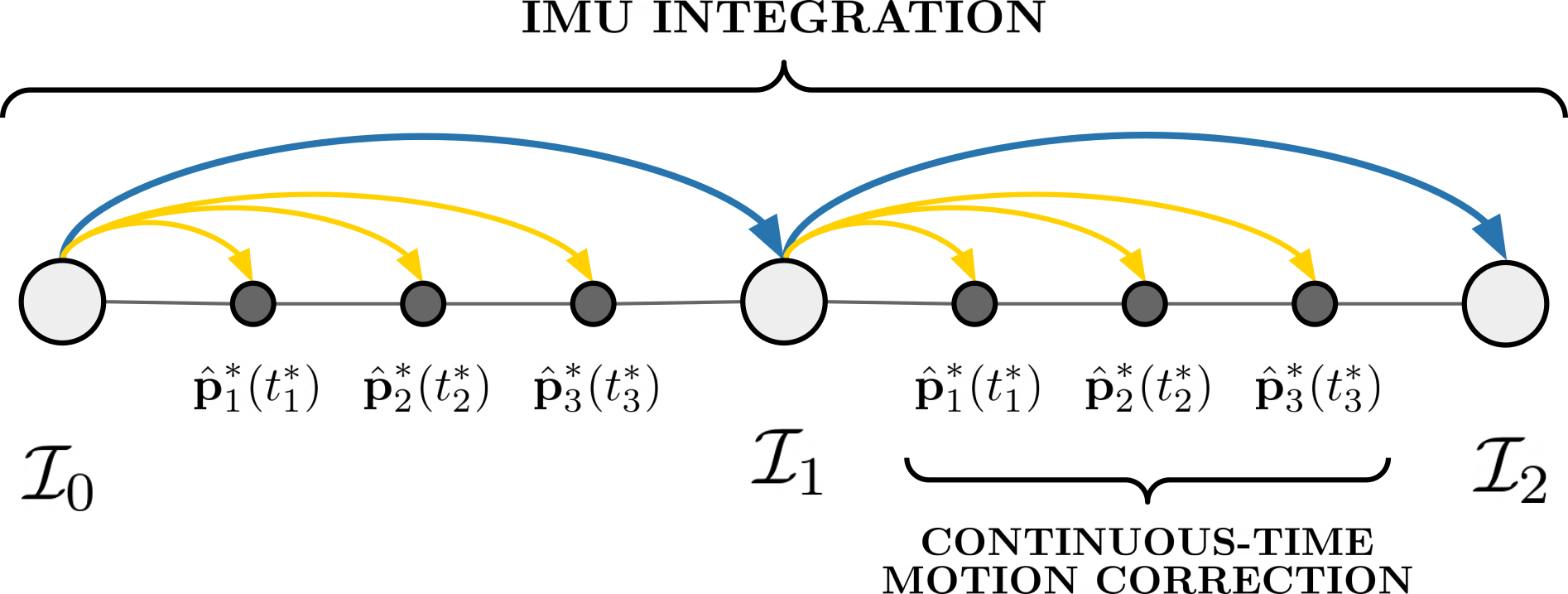}
    \vspace{-2mm}
    \caption{\textbf{Continuous-Time Motion Correction.} For each point in a cloud, a unique transform is computed by solving a set of closed-form motion equations initialized at the closest preceeding IMU measurement. This provides accurate and parallelizable continuous-time motion correction.}
    \label{fig:continuous_time}
\end{figure}

Let $t_{k}$ be the clock time of the received point cloud $\mathcal{P}_k^{\mathcal{R}}$ with $N$ number of accumulated points within the time period, and let $t_{k} + \Delta t_{k}^n$ be the timestamp of a point $p_{k}^n$ in the cloud. To approximate each point's location in $\mathcal{W}$, we first integrate IMU measurements between $t_{k\text{-}1}$ and $t_{k} + \Delta t_{k}^N$ via
\begin{equation}
\begin{alignedat}{2}
    \label{eq:deskew}
    \hat{\textbf{p}}_{i} &= \hat{\textbf{p}}_{i\text{-}1} &&+ \hat{\textbf{v}}_{i\text{-}1}\Delta t_i + \tfrac{1}{2}\hat{\textbf{R}}(\hat{\textbf{q}}_{i\text{-}1})\hat{\boldsymbol{a}}_{i\text{-}1}\Delta t_i^2 + \tfrac{1}{6}\hat{\boldsymbol{j}}_i\Delta t_i^3 \,,\\
    \hat{\textbf{v}}_{i} &= \hat{\textbf{v}}_{i\text{-}1} &&+ \hat{\textbf{R}}(\hat{\textbf{q}}_{i\text{-}1})\hat{\boldsymbol{a}}_{i\text{-}1}\Delta t_i \,,\\
    \hat{\textbf{q}}_{i} &= \hat{\textbf{q}}_{i\text{-}1} &&+ \tfrac{1}{2} (\hat{\textbf{q}}_{i\text{-}1} \otimes \hat{\boldsymbol{\omega}}_{i\text{-}1} )\Delta t_i + \tfrac{1}{4}(\hat{\textbf{q}}_{i\text{-}1} \otimes \hat{\boldsymbol{\alpha}_i}) \Delta t_i^2\,,
\end{alignedat}
\end{equation}
\noindent for $i = 1,\hdots,M$ for $M$ number of IMU measurements between two scans, where $\hat{\boldsymbol{j}}_i = \tfrac{1}{\Delta t_i} {(\hat{\textbf{R}}(\hat{\textbf{q}}_{i})\hat{\boldsymbol{a}}_i - \hat{\textbf{R}}(\hat{\textbf{q}}_{i\text{-}1})\hat{\boldsymbol{a}}_{i\text{-}1})}$ and $\hat{\boldsymbol{\alpha}}_i =  \tfrac{1}{\Delta t_i} {(\hat{\boldsymbol{\omega}}_i - \hat{\boldsymbol{\omega}}_{i\text{-}1})}$ are the estimated linear jerk and angular acceleration, respectively. The set of homogeneous transformations $\hat{\textbf{T}}_i^{\mathcal{W}} \in \mathbb{SE}(3)$ that correspond to $\hat{\textbf{p}}_i$ and $\hat{\textbf{q}}_i$ then define the coarse, \textit{discrete}-time trajectory during a sweep. Then, an analytical, \textit{continuous}-time solution from the nearest preceding transformation to each point $p_k^n$ recovers the point-specific deskewing transform $\hat{\textbf{T}}^{\mathcal{W}*}_{n}$, such that
\begin{equation}
\begin{alignedat}{2}
    \label{eq:deskew_timestamp}
        \hat{\textbf{p}}^*(t) &= \hat{\textbf{p}}_{i\text{-}1} &&+ \hat{\textbf{v}}_{i\text{-}1} t + \tfrac{1}{2}\hat{\textbf{R}}(\hat{\textbf{q}}_{i\text{-}1})\hat{\boldsymbol{a}}_{i\text{-}1} t^2 + \tfrac{1}{6}\hat{\boldsymbol{j}}_i t^3 \,,\\
        \hat{\textbf{q}}^*(t) &= \hat{\textbf{q}}_{i\text{-}1} &&+ \tfrac{1}{2} (\hat{\textbf{q}}_{i\text{-}1} \otimes \hat{\boldsymbol{\omega}}_{i\text{-}1}) t + \tfrac{1}{4} (\hat{\textbf{q}}_{i\text{-}1} \otimes \hat{\boldsymbol{\alpha}_i}) t^2 \,,
\end{alignedat}
\end{equation}
\noindent where ${i{-}1}$ and $i$ correspond to the closest preceding and successive IMU measurements, respectively, $t$ is the timestamp between point $p_k^n$ and the closest preceding IMU, and $\hat{\textbf{T}}^{\mathcal{W}*}_{n}$ is the transformation corresponding to $\hat{\textbf{p}}^*$ and $\hat{\textbf{q}}^*$ for $p_k^n$ (Fig.~\ref{fig:continuous_time}). Note that (\ref{eq:deskew_timestamp}) is parameterized only by $t$ and therefore a transform can be queried for any desired time to construct a continuous-time trajectory.

The result of this two-step procedure is a motion-corrected point cloud that is also approximately aligned with the map in $\mathcal{W}$, which therefore inherently incorporates the optimization prior used for GICP (Sec.~\ref{s2m}). Importantly, (\ref{eq:deskew}) and (\ref{eq:deskew_timestamp}) depend on the accuracy of $\hat{\textbf{v}}^{\mathcal{W}}_0$, the initial estimate of velocity, $\textbf{b}_k^a$ and $\textbf{b}_k^\omega$, the estimated IMU biases, in addition to an accurate initial body orientation $\hat{\textbf{q}}_0$ (to properly compensate for the gravity vector) at the time of motion correction. We therefore emphasize that, a key to the reliability of our approach is the \textit{guaranteed global convergence} of these terms by leveraging DLIO's nonlinear geometric observer \cite{lopez2023contracting}, provided that scan-matching returns an accurate solution.

\begin{table*}[!t]
    \centering
    \footnotesize
    \setlength{\tabcolsep}{4 pt}
    \renewcommand{\arraystretch}{1.2}
    \vspace{2mm}
    \caption{Comparison with Newer College Dataset}
    \vspace{-2mm}
    \begin{tabular}{|l|c||c|c|c|c|c|c|}
    \hline
   \multicolumn{1}{|c|}{\multirow{2}{*}{Algorithm}} 
    & \multicolumn{1}{c||}{\multirow{2}{*}{Type}} 
    & \multicolumn{5}{c|}{\multirow{1}{*}{Absolute Trajectory Error (RMSE) [m]}} 
    & \multicolumn{1}{c|}{\multirow{2}{*}{Avg Comp. [ms]}} \\ \cline{3-7}
    & & Short ($1609.40$m) & Long ($3063.42$m) & Quad ($479.04$m) & Dynamic ($97.20$m) & Park ($695.68$m) & \\ \hline
DLO \cite{chen2022direct} & LO & 0.4633 & 0.4125 & 0.1059 & 0.1954 & 0.1846 & 48.10\\ \hdashline
CT-ICP~\cite{dellenbach2022ct} & LO & 0.5552 & 0.5761 & 0.0981 & 0.1426 & 0.1802 & 412.27 \\ \hdashline
LIO-SAM~\cite{shan2020lio} & LIO & 0.3957 & 0.4092 & 0.0950 & 0.0973 & 0.1761 & 179.33 \\ \hdashline
FAST-LIO2~\cite{xu2022fast} & LIO & 0.3775 & 0.3324 & 0.0879 & 0.0771 & 0.1483 & 42.86 \\ \hdashline
DLIO (None) & LIO & 0.4299 & 0.3988 & 0.1117 & 0.1959 & 0.1821 & 34.88 \\ \hdashline
DLIO (Discrete) & LIO & 0.3803 & 0.3629 & 0.0943 & 0.0798 & 0.1537 & \textbf{34.61} \\ \hdashline
DLIO (Continuous) & LIO & \textbf{0.3606} & \textbf{0.3268} & \textbf{0.0837} & \textbf{0.0612} & \textbf{0.1196} & 35.74 \\ \hline
    \end{tabular}
    \vspace{-4mm}
    \label{table:newer_test}
\end{table*}

\subsection{Scan-to-Map Registration}
\label{s2m}

By simultaneously correcting for motion distortion and incorporating the GICP optimization prior into the point cloud, DLIO can directly perform scan-to-map registration and bypass the scan-to-scan procedure required in previous methods. This registration is cast as a nonlinear optimization problem which minimizes the distance of corresponding points/planes between the current scan and an extracted submap. Let $\hat{\mathcal{P}}_k^{\mathcal{W}}$ be the corrected cloud in $\mathcal{W}$ and $\hat{\mathcal{S}}_k^{\mathcal{W}}$ be the extracted keyframe-based submap via \cite{chen2022direct}. Then, the objective of scan-to-map optimization is to find a transformation $\Delta \hat{\textbf{T}}_k$ which better aligns the point cloud such that
\begin{equation}
    \label{eq:gicp}
    \Delta \hat{\textbf{T}}_k = \argmin_{\Delta \textbf{T}_k} \, \mathcal{E} \left( \Delta \textbf{T}_k \hat{\mathcal{P}}_{k}^{\mathcal{W}},\, \hat{\mathcal{S}}_{k}^{\mathcal{W}} \right) \,,
\end{equation}
\noindent where the GICP residual error $\mathcal{E}$ is defined as
\begin{equation*}
\vspace{-2mm}
    \mathcal{E} \left( \Delta \textbf{T}_k \hat{\mathcal{P}}_{k}^{\mathcal{W}}, \hat{\mathcal{S}}_{k}^{\mathcal{W}} \right) = \sum_{c \in \mathcal{C}} d_c^\top \left( C_{k,c}^{\mathcal{S}} + \Delta \textbf{T}_k C_{k,c}^{\mathcal{P}} \Delta \textbf{T}^\top_k \right)^{-1} d_c \,,
\end{equation*}
\noindent for a set of $\mathcal{C}$ corresponding points between $\hat{\mathcal{P}}_k^{\mathcal{W}}$ and $\hat{\mathcal{S}}_k^{\mathcal{W}}$ at timestep $k$,\, $d_c = \hat{s}_k^c - \Delta \textbf{T}_k \hat{p}_k^c$,\, $\hat{p}_k^c \in \hat{\mathcal{P}}_{k}^{\mathcal{W}}$,\, $\hat{s}_k^c \in \hat{\mathcal{S}}_{k}^{\mathcal{W}}$,\, $\forall c \in \mathcal{C}$, and $C_{k,c}^{\mathcal{P}}$ and $C_{k,c}^{\mathcal{S}}$ are the estimated covariance matrices for point cloud $\hat{\mathcal{P}}_k^{\mathcal{W}}$ and submap $\hat{\mathcal{S}}_k^{\mathcal{W}}$, respectively. Then, following \cite{segal2009generalized}, this point-to-plane formulation is converted into a plane-to-plane optimization by regularizing covariance matrices $C_{k,c}^{\mathcal{P}}$ and $C_{k,c}^{\mathcal{S}}$ with $(1, 1, \epsilon)$ eigenvalues, where $\epsilon$ represents the low uncertainty in the surface normal direction. 
The resulting $\Delta \hat{\textbf{T}}_k$ represents an optimal correction transform which better globally aligns the prior-transformed scan $\hat{\mathcal{P}}_k^{\mathcal{W}}$ to the submap $\hat{\mathcal{S}}_k^{\mathcal{W}}$, so that $\hat{\textbf{T}}^{\mathcal{W}}_k = \Delta \hat{\textbf{T}}_k \hat{\textbf{T}}_M^\mathcal{W}$ (where $\hat{\textbf{T}}_M^\mathcal{W}$ is the last point's IMU integration) is the globally-refined robot pose which is used for map construction and as the update signal for the nonlinear geometric observer.

\subsection{Geometric Observer}
\label{methods:geo}

The transformation $\hat{\textbf{T}}^{\mathcal{W}}_k$ computed by scan-to-map alignment is fused with IMU measurements to generate a full state estimate $\hat{\textbf{X}}_k$ via a novel hierarchical nonlinear geometric observer.  
A full analysis of the observer can be found in \cite{lopez2023contracting}, but in summary, one can show that $\hat{\textbf{X}}$ will globally converge to $\textbf{X}$ in the deterministic setting with minimal computation.
The proof utilizes contraction theory to first prove that the quaternion estimate converges exponentially to a region near the true quaternion.
The orientation estimate then serves as an input to another contracting observer that estimates translation states.
This architecture forms a contracting hierarchy that guarantees the estimates converge to their true values.
This strong convergence result is the main advantage over other fusion schemes, e.g., filtering or pose graph optimization, which possess minimal convergence guarantees even in the most ideal setting.
Additionally, the inherent smoothness of the observer's state estimate makes it suitable for control.
The observer used in this work is a special case of the one in \cite{lopez2023contracting}.

\begin{figure}[!t]
    \centering
    \vspace{-1mm}
    \includegraphics[width=0.95\columnwidth]{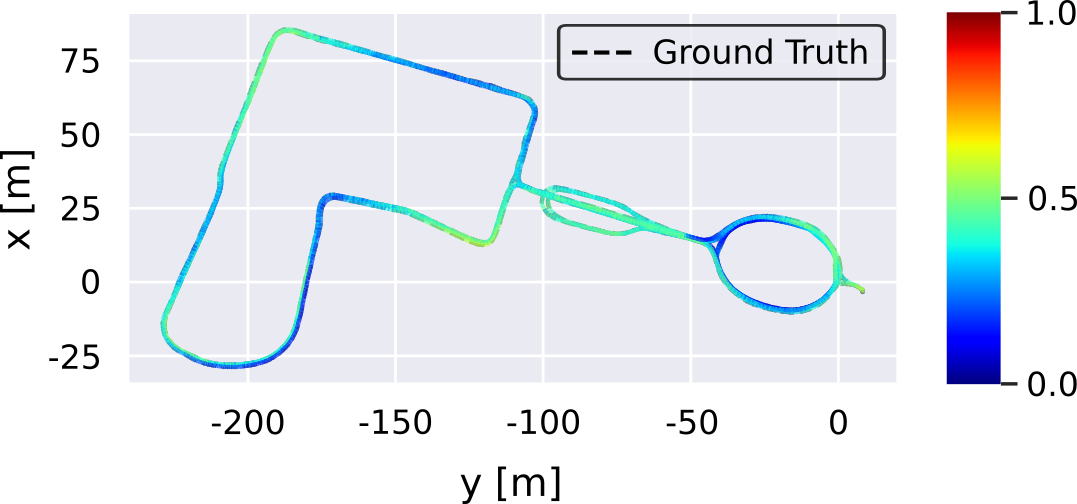}
    \vspace{-2mm}
    \caption{\textbf{Trajectory of Long Experiment.} DLIO's generated trajectory for the Newer College - Long Experiment. Color indicates absolute pose error.}
    \label{fig:trajectory}
\end{figure}

Let $\gamma_{\ell\in\{1,\dots,5\}}$ be positive constants and $\Delta t^{+}_k$ be the time between GICP poses.
If $\textbf{q}_e \coloneqq (q_e^\circ,~\vec{q}_e) = \hat{\textbf{q}}^*_i \otimes \hat{\textbf{q}}_k$ and $\textbf{p}_e = \hat{\textbf{p}}_k - \hat{\textbf{p}}_i$ (errors between propagated and measured poses) then the state correction takes the form
\begin{equation}
    \begin{alignedat}{3}
    \label{eq:update}
      &\hat{\textbf{q}}_i && \leftarrow \hat{\textbf{q}}_i &&+ \Delta t^{+}_k \, \gamma_1 \, \hat{\textbf{q}}_i \otimes \left[ \begin{array}{c} 1 - |q_e^\circ| \\ \mathrm{sgn}(q_e^\circ) \, \vec{q}_e \end{array} \right] \,,\\
      &\hat{\textbf{b}}^{\boldsymbol{\omega}}_i && \leftarrow \hat{\textbf{b}}^{\boldsymbol{\omega}}_i &&- \Delta t^{+}_k \, \gamma_2 \, q_e^\circ \vec{q}_e  \,,\\
      &\hat{\textbf{p}}_i && \leftarrow \hat{\textbf{p}}_i &&+ \Delta t^{+}_k \, \gamma_3 \, \textbf{p}_e  \,,\\
      &\hat{\textbf{v}}_i && \leftarrow \hat{\textbf{v}}_i &&+ \Delta t^{+}_k \, \gamma_4 \, \textbf{p}_e  \,,\\
      &\hat{\textbf{b}}^{\boldsymbol{a}}_i && \leftarrow \hat{\textbf{b}}^{\boldsymbol{a}}_i &&- \Delta t^{+}_k \, \gamma_5 \, \hat{\textbf{R}}(\hat{\textbf{q}}_i)^\top \textbf{p}_e \,.\\
    \end{alignedat}
\end{equation}
Note (\ref{eq:update}) is hierarchical as the attitude update (first two eqs.) is completely decoupled from the translation update (last three eqs.).
Also, (\ref{eq:update}) is a fully nonlinear update which allows one to guarantee the state estimates are accurate enough to directly perform scan-to-map registration solely with an IMU prior without the need for scan-to-scan.


\begin{table*}[!t]
    \centering
    \footnotesize
    \setlength{\tabcolsep}{8 pt}
    \renewcommand{\arraystretch}{1.15}
    \vspace{2mm}
    \caption{Comparison with UCLA Campus Dataset}
    \vspace{-2mm}
    \begin{tabular}{|l|c||c|c|c|c|c|c|c|c|}
        \hline
        \multicolumn{1}{|c|}{\multirow{2}{*}{Algorithm}} &
        \multicolumn{1}{c||}{\multirow{2}{*}{Type}} &
          \multicolumn{4}{c|}{End-to-End Translational Error {[}m{]}} &
          \multicolumn{4}{c|}{Avg. Comp. {[}ms{]}} \\ \cline{3-10} & &
          \multicolumn{1}{c|}{A ($652.66$m)} &
          \multicolumn{1}{c|}{B ($526.58$m)} &
          \multicolumn{1}{c|}{C ($551.38$m)} &
          \multicolumn{1}{c|}{D ($530.75$m)} &
          \multicolumn{1}{c|}{A} &
          \multicolumn{1}{c|}{B} &
          \multicolumn{1}{c|}{C} &
          \multicolumn{1}{c|}{D} \\ \hline
        DLO~\cite{chen2022direct} & LO & 0.0216 & 1.2932 & 0.0375 & 0.0178 & 20.40 & 20.77 & 21.18 & 21.62 \\ \hdashline
        CT-ICP~\cite{dellenbach2022ct} & LO & 0.0387 & 0.0699 & 0.0966 & 0.0253 & 351.85 & 342.76 & 334.15 & 370.19 \\ \hdashline
        LIO-SAM~\cite{shan2020lio} & LIO & 0.0216 & 0.0692 & 0.0936 & 0.0249 & 33.21 & 29.14 & 39.04 & 48.94 \\ \hdashline
        FAST-LIO2~\cite{xu2022fast} & LIO & 0.0454 & 0.0353 & 0.0363 & 0.0229 & 15.39 & 12.25 & 14.84 & 15.01 \\ \hdashline
        DLIO & LIO & \textbf{0.0105} & \textbf{0.0233} & \textbf{0.0301} & \textbf{0.0082} & \textbf{10.45} & \textbf{8.37} & \textbf{8.66} & \textbf{10.96} \\ \hline
    \end{tabular}
    \vspace{-2mm}
    \label{table:ucla}
\end{table*}

\begin{figure}[!t]
    \centering
    \vspace{-1mm}
    \includegraphics[width=0.95\columnwidth]{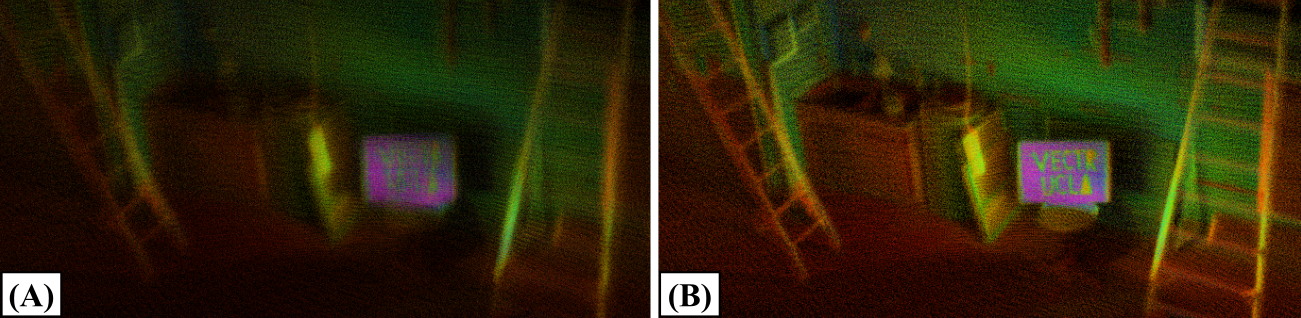}
    \vspace{-2mm}
    \caption{\textbf{Deskewing Comparison.} Map generated from aggressive maneuvers without (A) and with (B) our motion correction method.}
    \label{fig:deskew_comparison}
\end{figure}

\begin{figure*}[!t]
    \centering
    \includegraphics[width=0.99\textwidth]{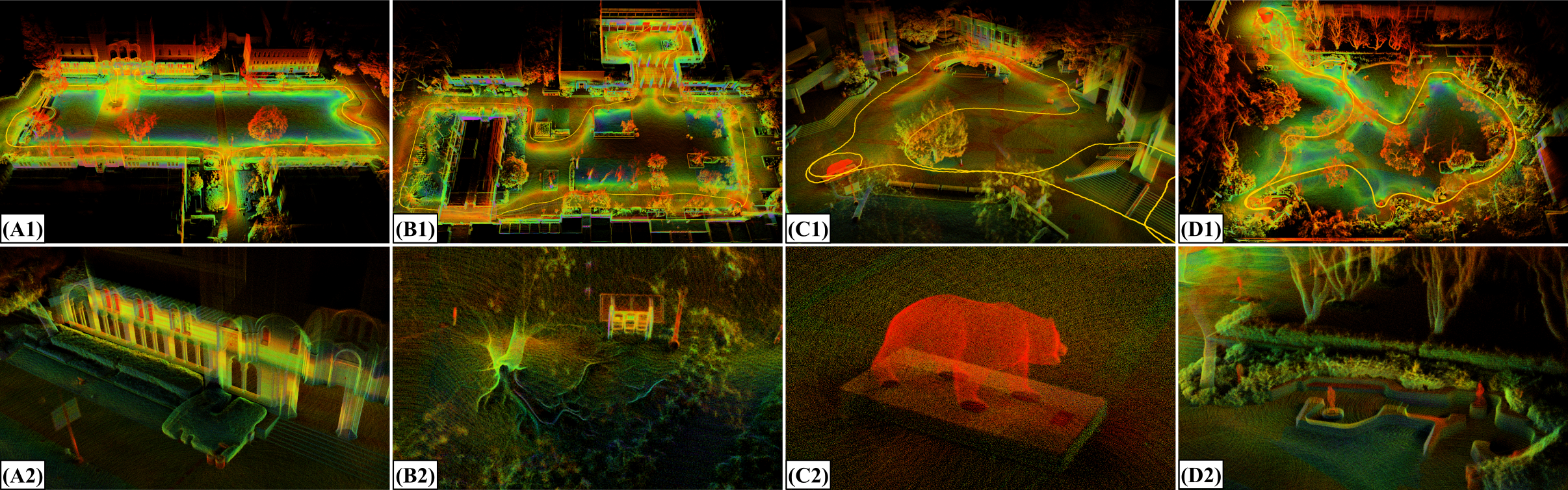}
    \vspace{-2mm}
    \caption{\textbf{UCLA Campus.} Detailed maps of locations around UCLA in Los Angeles, CA generated by DLIO, including (A) Royce Hall in Dickson Court, (B) Court of Sciences, (C) Bruin Plaza, and (D) the Franklin D. Murphy Sculpture Garden, with both (1) a bird's eye view and (2) a close-up to demonstrate the level of fine detail DLIO can generate. The trajectory taken to generate these maps is shown in yellow in the first row.}
    \vspace{-6mm}
    \label{fig:ucla_maps}
\end{figure*}

\section{Results}

DLIO was evaluated using the Newer College benchmark dataset~\cite{ramezani2020newer} and data self-collected around the UCLA campus. We compare accuracy and efficiency against four state-of-the-algorithms, namely DLO~\cite{chen2022direct}, CT-ICP~\cite{dellenbach2022ct}, LIO-SAM~\cite{shan2020lio}, and FAST-LIO2~\cite{xu2022fast}. Each algorithm employs a different degree and method of motion compensation, therefore creating an exhaustive comparison to the current state-of-the-art. Aside from extrinsics, default parameters at the time of writing for each algorithm were used in all experiments unless otherwise noted. Specifically, loop-closures were enabled for LIO-SAM and online extrinsics estimation disabled for FAST-LIO2 to provide the best results of each algorithm. For CT-ICP, voxelization was slightly increased and data playback was slowed down otherwise the algorithm would fail due to significant frame drops. All tests were conducted on a 16-core Intel i7-11800H CPU.

\subsection{Ablation Study and Comparison of Motion Correction}

To investigate the impact of our proposed motion correction scheme, we first conducted an ablation study with varying degrees of deskewing in DLIO using the Newer College dataset~\cite{ramezani2020newer}. This study ranged from no motion correction (None), to correction using only nearest IMU integration via (\ref {eq:deskew}) (Discrete), and finally to full continuous-time motion correction via both (\ref{eq:deskew}) and (\ref{eq:deskew_timestamp}) (Continuous) (Table~\ref{table:newer_test}). Particularly of note is the Dynamic dataset, which contained highly aggressive motions with rotational speeds up to 3.5 rad/s. With no correction, error was the highest among all algorithms at $0.1959$ RMSE. With partial correction, error significantly reduced due to scan-matching with more accurate and representative point clouds; however, using the full proposed scheme, we observed an error of only $0.0612$ RMSE---the lowest among all tested algorithms. With similar trends for all other datasets, the superior tracking accuracy granted by better motion correction is clear: constructing a unique transform in continuous-time creates a more authentic point cloud than previous methods, which ultimately affects scan-matching and therefore trajectory accuracy. Fig.~\ref{fig:deskew_comparison} showcases this empirically: DLIO can capture minute detail that is otherwise lost with simple or no motion correction.

\subsection{Benchmark Results}

\subsubsection{Newer College Dataset}

Trajectory accuracy and average per-scan time of all algorithms were also compared using the original Newer College benchmark dataset~\cite{ramezani2020newer} via evo~\cite{grupp2017evo}. For these tests, we used data from the Ouster's IMU (100Hz) alongside LiDAR data (10Hz) to ensure accurate time synchronization between sensors. For certain Newer College datasets, the first 100 poses were excluded from computing FAST-LIO2's RMSE due to slippage at the start in order to provide a fair comparison. We also compared using the recent extension of the Newer College dataset~\cite{zhang2021multicamera} and observed similar results, but those results have been omitted due to space constraints. The results are shown in Table~\ref{table:newer_test}, in which we observed DLIO to produce the lowest trajectory RMSE and lowest overall per-scan computational time (averaged across all five datasets) as compared to the state-of-the-art.
Fig.~\ref{fig:trajectory} illustrates DLIO's low trajectory error compared to ground truth for the Newer College - Long Experiment dataset even after over three kilometers of travel.

\subsubsection{UCLA Campus Dataset}

We additionally collected four large-scale datasets at UCLA for additional comparison (Fig.~\ref{fig:ucla_maps}). These datasets were gathered by hand-carrying our aerial platform (Fig.~\ref{fig:main}) over 2261.37m of total trajectory. Our sensor suite included an Ouster OS1 (10Hz, 32 channels recorded with a 512 horizontal resolution) and a 6-axis InvenSense MPU-6050 IMU located approximately 0.1m below it. We note here that this IMU can be purchased for approximately \$10, demonstrating that LIO algorithms need not require high-grade IMU sensors that previous works have used. Note that a comparison of absolute trajectory error was not possible due to the absence of ground truth, so as is common practice, we compute end-to-end translational error as a proxy metric (Table~\ref{table:ucla}). In these experiments, DLIO outperformed all others across the board in both end-to-end translational error and per-scan efficiency. DLIO's resulting maps can capture fine detail in the environment which ultimately provides more intricate information cues for autonomous mobile robots such as terrain traversability.


\section{Conclusion}

This work presents Direct LiDAR-Inertial Odometry (DLIO), a highly reliable LIO algorithm that yields accurate state estimates and detailed maps in real-time for resource-contrained mobile robots. The key innovation that distinguishes DLIO from others is its fast and parallelizable coarse-to-fine approach in constructing continuous-time trajectories for point-wise motion correction. This approach is built into a simplified LIO architecture which performs motion correction and prior construction in one shot, in addition directly performing scan-to-map alignment for reduced computational overhead. This is all feasible due to our observer's strong convergence guarantees which reliably initializes pose, velocity, and biases for accurate IMU integration. Our experimental results demonstrate DLIO's improved localization accuracy, map clarity, and algorithmic efficiency as compared to the state-of-the-art, and future work includes closed-loop flight tests and adding loop closures.


\vspace{4mm}
\noindent \small \textbf{Acknowledgements:} The authors would like to thank Helene Levy and David Thorne for their help with data collection.


\newpage
\bibliographystyle{IEEEtran}
\bibliography{references}

\end{document}